\documentclass[journal]{IEEEtran}

\ifCLASSINFOpdf
\else
   \usepackage[dvips]{graphicx}
\fi
\usepackage{url}

\usepackage[mathletters]{ucs}
\usepackage[utf8x]{inputenc}

\usepackage{float}
\usepackage{graphicx, cite}
\usepackage{multirow, booktabs, makecell}
\usepackage[table]{xcolor}
\definecolor{gray}{rgb}{0.85,0.85,0.85}
\usepackage{footnote}
\makesavenoteenv{table}

\usepackage{amsmath}
\usepackage{amsfonts}
\usepackage{amssymb}
\usepackage{amsthm}
\usepackage{braket}
\usepackage{nicematrix}
\usepackage[scientific-notation=true]{siunitx}

\usepackage{mathtools}
\usepackage{bbold}
\usepackage{algorithm} 
\usepackage{algpseudocode, setspace}
\usepackage{siunitx}
\usepackage{graphicx,float}
\usepackage[table]{xcolor}
\definecolor{gray}{rgb}{0.85,0.85,0.85}
\usepackage{tabularray}
\usepackage{booktabs,arydshln}
\usepackage{xcolor}
\usepackage[export]{adjustbox}
\usepackage{svg}
\usepackage{multicol,multirow}
\usepackage{subcaption}

\usepackage{hyperref}
\hypersetup{
	colorlinks   = true,    
	urlcolor     = blue,    
	linkcolor    = red,     
	citecolor    = green    
}

\graphicspath{{IMAGES/}}
\definecolor{dg}{rgb}{0.0, 0.5, 0.0}

\begin{document}

\title{Facial Image Feature Analysis and its Specialization for Fr\'echet Distance and Neighborhoods}

\author{Doruk Cetin$^{*\ddagger}$ \qquad Benedikt Schesch$^{*\dagger}$ \qquad Petar Stamenkovic$^{\dagger}$ \\ Niko Benjamin Huber$^{\ddagger}$ \qquad Fabio Z\"und$^{\dagger}$ \qquad Majed El Helou$^{\dagger}$

\thanks{$^*$ Equal contribution, second author was an intern at MTC.}
\thanks{$^{\ddagger}$Align Technology Z\"urich, Switzerland. $^{\dagger}$Media Technology Center, ETH Z\"urich, Switzerland. dcetin@aligntech.com, \{bschesch, pstamenkovic, melhelou\}@ethz.ch.}
}

\maketitle
\IEEEpeerreviewmaketitle

\begin{abstract}
Assessing distances between images and image datasets is a fundamental task in vision-based research. It is a challenging open problem in the literature and despite the criticism it receives, the most ubiquitous method remains the Fr\'echet Inception Distance. The Inception network is trained on a specific labeled dataset, ImageNet, which has caused the core of its criticism in the most recent research. Improvements were shown by moving to self-supervision learning over ImageNet, leaving the training data domain as an open question. We make that last leap and provide the first analysis on domain-specific feature training and its effects on feature distance, on the widely-researched facial image domain. We provide our findings and insights on this domain specialization for Fr\'echet distance and image neighborhoods, supported by extensive experiments and in-depth user studies.
\end{abstract}


\vspace{-0.4cm}

\section{Introduction and Related Work} \label{sec:intro}
Measuring distances between datasets is a valuable yet challenging task, in particular for complex signals such as images. It is crucial for understanding data distributions and domain gaps for transfer learning and generalization. It is also important for developing generative networks that recently gained in popularity~\cite{karras2020analyzing} and that are prone to hallucination~\cite{el2022bigprior}. 

The most ubiquitous approach to measuring dataset distance is the widely used Fr\'echet inception distance (FID)~\cite{heusel2017gans}. It computes the Fr\'echet statistical distance~\cite{frechet1957distance} between the datasets' image features, extracted by an ImageNet-trained~\cite{deng2009imagenet} Inception network~\cite{szegedy2016rethinking}. A plethora of similar solutions emerged in the literature, notably extensions to conditional inputs~\cite{soloveitchik2021conditional} and adversarial robustness~\cite{alfarra2022robustness}. KID~\cite{binkowski2018demystifying} proposes a modified distance on the same feature space. It has certain theoretical advantages but in practice correlates closely with FID. CKA~\cite{yang2023revisiting} is another FID alternative, but both the paper's results and user survey show it performs in a similar way as FID. sFID~\cite{nash2021generating} simply mimics FID on intermediate feature maps to improve spatial information that is more fine-grained. StyleGAN-XL~\cite{sauer2022styleganxl} even computes rFID on the features of a randomly initialized network as an additional metric, with the idea originating from Naeem~\emph{et al.}~\cite{naeem2020reliable}, where the objective is to be more general by being task-agnostic. rFID results, however, tend to have an erratic behavior in practice with extremely large values. Zhang~\emph{et al.} ~\cite{zhang2018unreasonable} have shown that training is crucial as random networks achieve significantly worse performance, and that random networks focus more on low-level information~\cite{lee2023demystifying}, further supporting the use of trained features for Fr\'echet distance.

More specialized methods have been proposed that are different from FID. Kynk\"a\"anniemi~\emph{et al.}~\cite{kynkaanniemi2019improved} investigated precision and recall between feature spaces as complementary metrics to FID. The pair can illustrate certain trade-offs but do not give a single score that can be used as an optimization target. We note that the experimental evaluation on faces only uses the standard ImageNet FID~\cite{kynkaanniemi2019improved}. Precision and recall definitions are refined to density and coverage in~\cite{naeem2020reliable}, to better adapt to image manifolds that need to be estimated through only a limited number of sample points. Another approach to visualize feature space drift is based on SVCCA~\cite{el2020al2}, however, it cannot readily scale to large datasets. Lastly and most similar to FID, Ramtoula~\emph{et al.}~\cite{ramtoula2023visual} create a histogram per neuron that contains the activation values of that neuron across network layers. The histogram of an image can be compared to the average histogram of a dataset to obtain a similarity metric. While this approach has the advantage of applicability to a single image, high-level information in cross-neuron dependencies as well as location information are lost. 

FID thus remains the most practical and ubiquitous metric in recent literature~\cite{parmar2022aliased,lee2023demystifying}, despite its numerous shortcomings. The most simple to resolve are that it can be affected by resizing when anti-aliasing is omitted~\cite{parmar2022aliased} and that its estimator has statistical bias that is model dependent~\cite{chong2020effectively}. FID relies on an ImageNet-trained Inception network that can be more sensitive to texture than to shape~\cite{geirhos2018imagenet}. This bias is due to aggressive random cropping in data augmentation and can be reduced by using more natural augmentations like image distortions~\cite{hermann2020origins}. However, the underlying ImageNet training causes inherent limitations~\cite{betzalel2022study}, such as a bias towards only the most salient object in a multi-object image~\cite{van2020investigating}, because ImageNet is meant for single object learning. Most recently, ~\cite{kynkaanniemi2023role} criticizes the strong relation between Inception features and ImageNet classes, particularly as ImageNet does \textit{not} contain human or human face classes while FID is most commonly used in studying generative models for face synthesis. 

The goal of Morozov~\emph{et al.}~\cite{morozov2021self} is to explore replacing supervised ImageNet feature extractors with self-supervised ones. The results show certain improvements in FID. The investigation supports the use of self supervision and concludes with the open question of using self-supervised features that are \textit{domain specific}, which was left to future research~\cite{morozov2021self}.

Our goal is to analyze how specializing the feature space impacts feature distances. We collect a novel facial dataset for our self-supervised learning to guarantee the independence from public datasets, and high image quality. We conduct extensive experiments and three user studies with 3432 answers from 26 participants.

\section{Methodology} ~\label{sec:method}

\vspace{-0.7cm}

\subsection{Feature-learning independent dataset}~\label{subsec:dataset}
To train our feature extractor, it is important to rely on a completely external dataset. The reason is that common facial image datasets are often used in training image generators, and any distance metric should be disentangled from them. 
We thus collect an in-house facial image dataset to train our feature extractor through self-supervision. We create a 30'000 image training set, in accordance with the size of CelebA-HQ~\cite{CelebAMask-HQ}, which we call Faces. The images are all center cropped, with no occlusions, and manually curated to ensure quality. By training a feature extractor on our held-out dataset, we lay the basis for an independent metric built over those features. This enables benchmarking on the commonly used public datasets, and on public image generators trained on them. To promote better fairness, our dataset is balanced across six ethnicities (latino hispanic, asian, middle eastern, black, indian, and white)~\cite{karkkainen2021fairface}. We further leave out an additional 21'000 images that we label for gender and use as a test set in a separate experiment outlined in Sec.~\ref{subsec:MLP}.

\subsection{Self-supervised feature learning}
Self-supervised learning can improve feature extraction performance~\cite{morozov2021self}. Another advantage is to decrease biases and errors coming from the choice and assignment of labels for supervised learning~\cite{kynkaanniemi2023role}. We exploit the simple yet effective state-of-the-art DINO~\cite{caron2021emerging} method for self-supervised learning on our dataset. DINO builds on knowledge distillation between teacher and student networks, and fundamental self-supervised learning data augmentation strategies, notably extending on SwAV~\cite{caron2020unsupervised}.
For all our experiments, we 
configure the feature embedding to have 2048 dimensions, aligning with the Inception~\cite{szegedy2016rethinking} architecture for direct comparisons. We train for 100 epochs on one 24GB NVIDIA RTX 3090 GPU with a batch size of 16, and all other settings follow DINO's approach.

\subsection{Fr\'echet distance over feature spaces}
The Fr\'echet~\cite{frechet1957distance} distance $F$ between two Gaussian distributions $\mathcal{N}(\mu_1,\Sigma_1)$ and $\mathcal{N}(\mu_2,\Sigma_2)$ is given by
\begin{equation*}
    F(\mu_1,\Sigma_1,\mu_2,\Sigma_2) = || \mu_1 - \mu_2 ||_2^2 + Tr(\Sigma_1 + \Sigma_2 -2 (\Sigma_1\Sigma_2)^\frac{1}{2} ),
\end{equation*}
where $Tr(\cdot)$ is the matrix trace. This formulation is then adapted to measure the distance between two datasets $\mathcal{D}_1$ and $\mathcal{D}_2$. This is achieved by exploiting the Inception~\cite{szegedy2016rethinking} network's feature extractor trained on ImageNet in a supervised manner, and called FID~\cite{heusel2017gans}. The feature extractor takes in an image and generates its embedding in a feature space. Generally, for any feature extractor $f(\cdot)$ we can define the Fr\'echet distance between datasets as $F(\mu^f_{\mathcal{D}_1},\Sigma^f_{\mathcal{D}_1},\mu^f_{\mathcal{D}_2},\Sigma^f_{\mathcal{D}_2})$, where $\mu^f_{\mathcal{D}_i}$ and $\Sigma^f_{\mathcal{D}_i}$ are the mean and covariance of the best-fit Gaussian over the feature distribution of dataset $i$, obtained by the feature extractor $f(\cdot)$. In our experiments, we study the effects of $f(\cdot)$ on the distance metric, with a focus on domain-specific specialized features. We denote the Fr\'echet distance computed over our DINO Faces feature space by FDD.

\begin{table}[t]
    \centering
    \caption{Classification accuracy (\%) of Head networks~\cite{caron2021emerging} and MLPs trained on CelebA-HQ features and the corresponding classes. CelebA-HQ features are extracted by: Inception (trained on ImageNet), DINO trained on ImageNet (I) and on Faces (F). We test on 3 \textcolor{red}{CelebA-HQ} classes and a \textcolor{blue}{Faces} class (gender) that we collected to have a fully separate test set.}
    \begin{NiceTabular}{l|c|c|c|c}
    \toprule
    Method /vs./ Test & \textcolor{red}{Blond} & \textcolor{red}{Young} & \textcolor{red}{Gender} & \textcolor{blue}{Gender} \\
    \midrule
    Inception + Head  & 93.54                  & \textbf{85.58}         & \textbf{96.44}          & 84.92 \\
    Inception + MLP   & 92.83                  & 83.90                  & 96.25                   & 84.22 \\
    \hdashline
    DINO (I) + Head   & 90.63                  & 83.08                  & 94.33                   & \textbf{86.40} \\
    DINO (I) + MLP    & 91.37                  & 83.25                  & 94.96                   & 85.71 \\
    \hdashline
    DINO (F) + Head   & \textbf{93.85}         & 82.54                  & 92.56                   & 85.86 \\
    DINO (F) + MLP    & 93.92                  & 83.06                  & 93.02                   & 86.00 \\ 
    \bottomrule
    \end{NiceTabular}
    \vspace{-0.4cm}
    \label{table:classification}
\end{table}

\begin{figure*}[t]
    \centering
    \includegraphics[width=.96\textwidth]{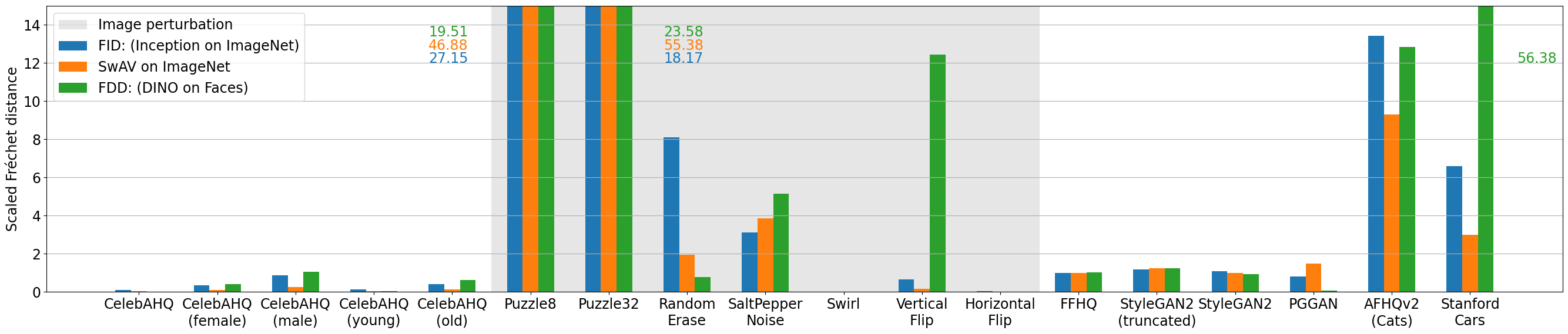}
    \vspace{-0.1cm}
    \caption{Rescaled Fr\'echet distances computed on Inception features (trained on ImageNet), SwAV features (trained on ImageNet), and DINO features (trained on Faces). Each distance is between the x-axis sets and CelebA-HQ data (5k samples). For better readability, we rescale all values with a ratio \textit{fixed per method} and determined on an independent dataset.}
    \label{fig:celebahq_fid_trained}
\end{figure*}

\section{Experimental Evaluation}~\label{sec:experiments}

\vspace{-0.6cm}

\subsection{Are our self-learned features sufficient?}~\label{subsec:MLP}
We evaluate whether our self-supervised features extract sufficient information relevant to faces. We train MLPs and Head networks on top of ImageNet-trained \textit{Inception} features (used by FID), \textit{DINO (I)} features from the ImageNet-trained DINO, and \textit{DINO (F)} features from the Faces-trained DINO. The MLPs and Heads are trained on the CelebA-HQ training set annotations to predict different binary classes (Blond, Young, Gender) based on input features. We show the results in Table~\ref{table:classification} on the CelebA-HQ test set and on an additional test set for gender from a fully independent source (Sec.~\ref{subsec:dataset}). We note that accuracies are significantly high, indicating that the networks extract sufficient features to enable classification. The results with the self-supervised DINO are on-par with Inception results, even surpassing them when testing on the independent curated test set, rather than the test set of CelebA-HQ. We emphasize, however, that the results highlight that the features are \textit{sufficient} but not that they are \textit{necessary}, in other words, some could nonetheless be irrelevant.

\subsection{Benchmarking Fr\'echet distance results}~\label{subsec:histogram}
We run benchmarking experiments for Fr\'echet distance computed over features from Inception (trained on ImageNet with supervision), SwAV (trained on ImageNet with self supervision), and DINO (trained on Faces with self supervision) networks. The distances are computed for 19 image sets (Fig.~\ref{fig:celebahq_fid_trained}) with respect to CelebA-HQ images, for 5k samples.

With DINO specialized to the facial domain with standardized faces, the distance is large when images are flipped vertically, while Inception and SwAV distances remain surprisingly small (smaller than the distance relative to FFHQ~\cite{karras2019style}, which also contains faces). For random erasing of small patches, the distance is the smallest for DINO, which can extract high-level facial features rather than only fine-granularity generalized ones, due to its specialization to faces. Meanwhile, Inception distance caused by random erasing is even larger than the Inception distance between Cars and CelebA-HQ. Lastly, we note the large distance for DINO on car images, which are completely out of domain. This is not the case with cats, where facial features remain correlated to human facial features and are aligned in the same way in preprocessing. For the remaining setups, the distances obtained by the different approaches remain, on average, closely tied.

\begin{figure}
    \centering
    \includegraphics[width=\linewidth]{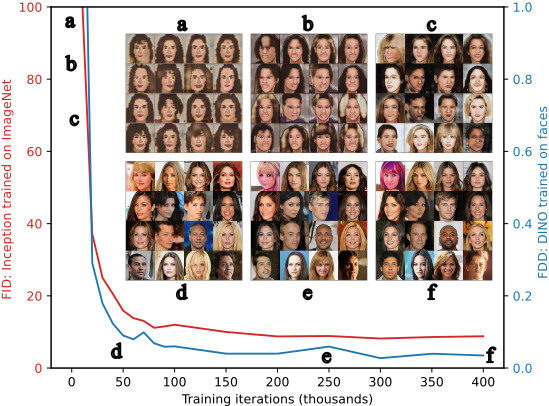}
    \caption{Fr\'echet distances computed on ImageNet-trained Inception features and on Faces-trained DINO features, between synthetically generated images and CelebA-HQ images.}
    \label{fig:anon_semantic}
\end{figure}

We also observe similar trends in Fig.~\ref{fig:anon_semantic} when tracking the training of the SemanticStyleGAN~\cite{shi2022semanticstylegan} with Fr\'echet distance on Inception and DINO. We only note that FDD is larger and drops faster than FID, as it is more sensitive to the low-quality faces initially synthesized. We  conduct a user study (with 10 images per class) to obtain ratings on how well an image corresponds to the CelebA-HQ distribution represented by randomly sampled sets of 9 images at a time (Table~\ref{table:study-correspondence}). While the variation in scores over classes such as male and female is interesting, we mostly note that FID and FDD (relative to CelebA-HQ) strongly correlated with the participants' answers (lower distance correlates with higher correspondence).

\begin{table}[t]
    \caption{User study rating how well images from different distributions correspond to random CelebA-HQ sets (1-5 score), the corresponding \textit{rescaled} Fr\'echet distances (FID, FDD), and Pearson and Spearman correlation.}
    \centering
    \begin{NiceTabular}{l|c|c|c|c}
    \toprule
    Image source distribution          & $\mu$ & $\sigma$ & FID & FDD \\
    \midrule
    CelebA-HQ (class: male)             & 2.00 & 1.09 & 0.87  & 1.06 \\
    CelebA-HQ (class: female)           & 2.52 & 1.15 & 0.34  & 0.40 \\
    CelebA-HQ (class: young)            & 2.43 & 1.20 & 0.12  & 0.06 \\
    CelebA-HQ (class: old)              & 2.28 & 1.16 & 0.43  & 0.63 \\ 
    StyleGAN2 (untruncated)            & 1.92 & 1.00 & {1.09} & {0.94} \\
    StyleGAN2 (0.7 truncated)          & 2.16 & 1.10 & 1.20  & {1.23} \\ \hdashline
    $r$-correlation to survey $\mu$    & 1.00 & -    & -0.83 & -0.79 \\ 
    $\rho$-correlation to survey $\mu$ & 1.00 & -    & -0.77 & -0.71 \\ 
    \bottomrule
    \end{NiceTabular}
    \vspace{-0.4cm}
    \label{table:study-correspondence}
\end{table}

\begin{figure*}[th]
        \centering
        \begin{subfigure}{.14\linewidth}
                \includegraphics[width=\linewidth]{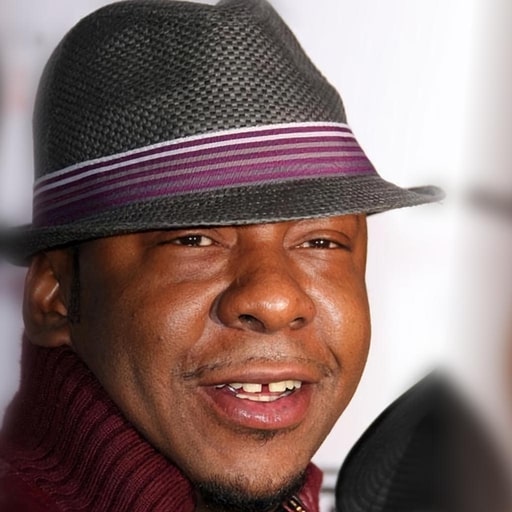}
        \end{subfigure}
        \begin{subfigure}{.42\linewidth}
                \includegraphics[width=\linewidth,trim={0 512 0 0},clip]{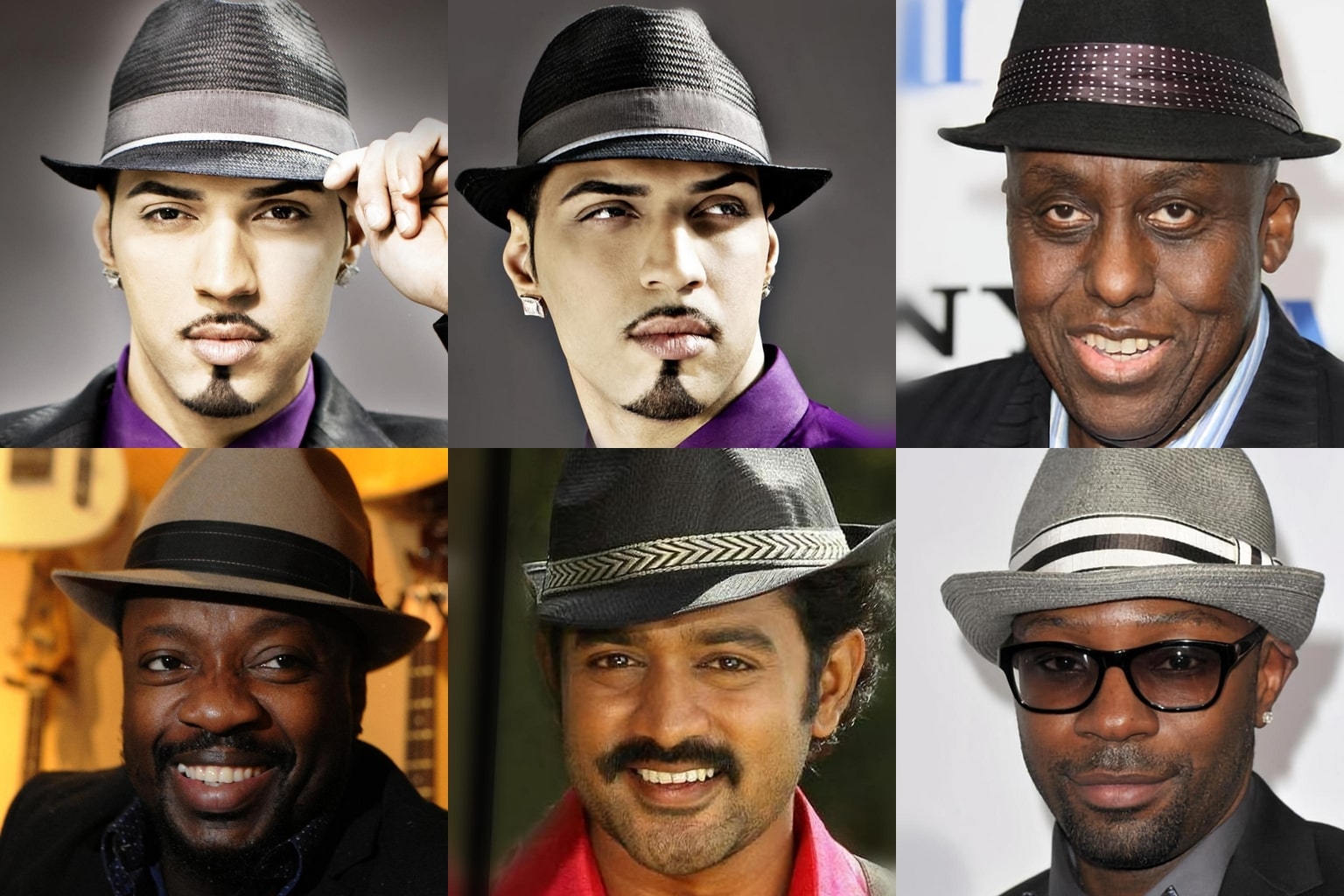}
        \end{subfigure}
        \begin{subfigure}{.42\linewidth}
                \includegraphics[width=\linewidth,trim={0 512 0 0},clip]{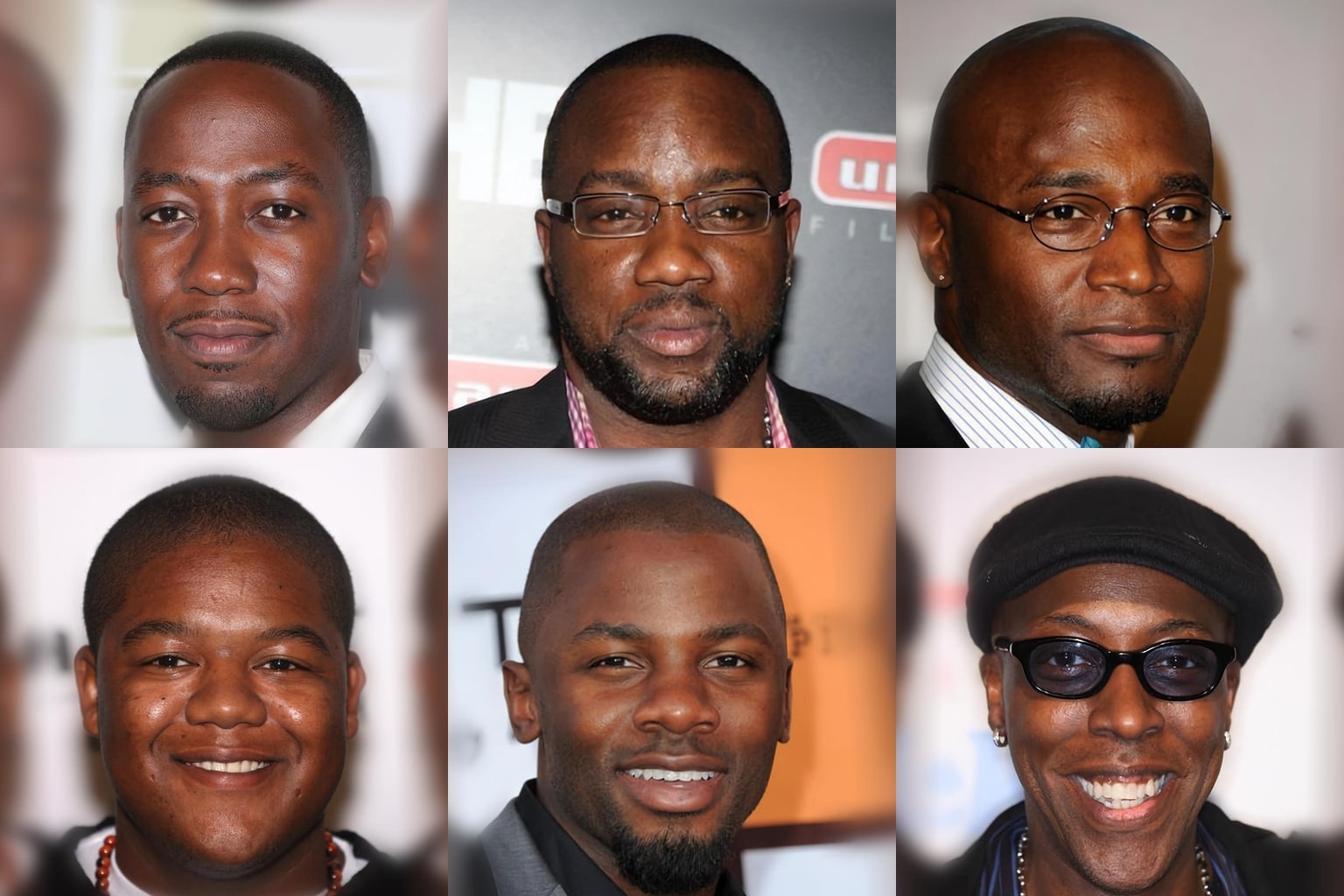}
        \end{subfigure}
        \begin{subfigure}{.14\linewidth}
                \includegraphics[width=\linewidth]{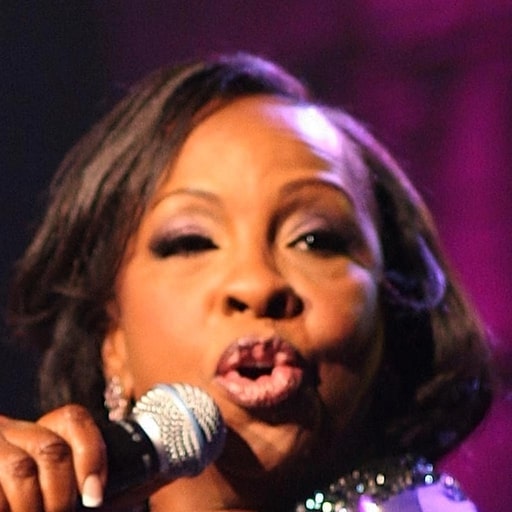}
                \caption{Reference}\label{fig:neighbors_00002_reference}
        \end{subfigure}
        \begin{subfigure}{.42\linewidth}
                \includegraphics[width=\linewidth,trim={0 512 0 0},clip]{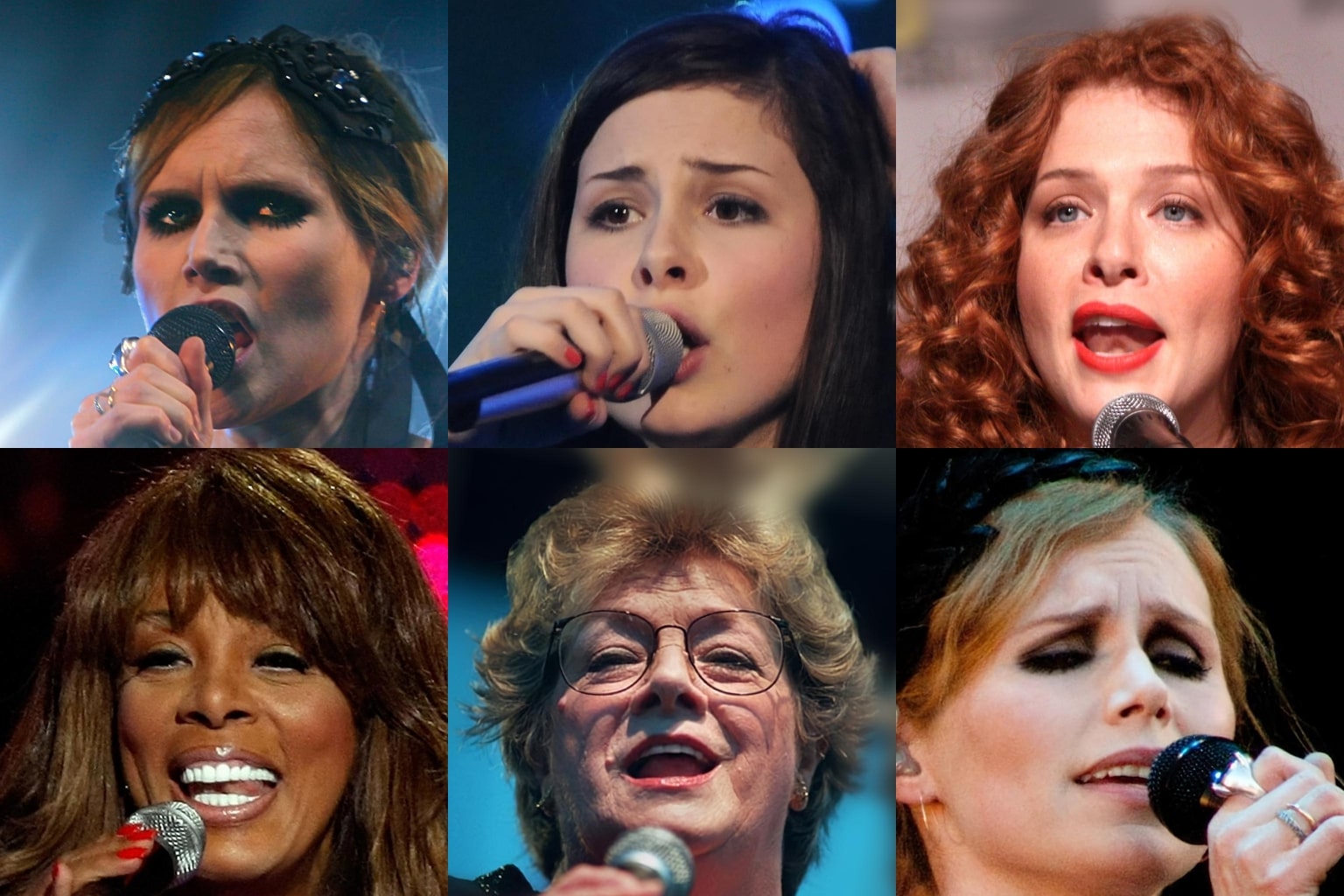}
                \caption{Nearest neighbors in the Inception space (FID)}\label{fig:neighbors_00002_inception}
        \end{subfigure}
        \begin{subfigure}{.42\linewidth}
                \includegraphics[width=\linewidth,trim={0 512 0 0},clip]{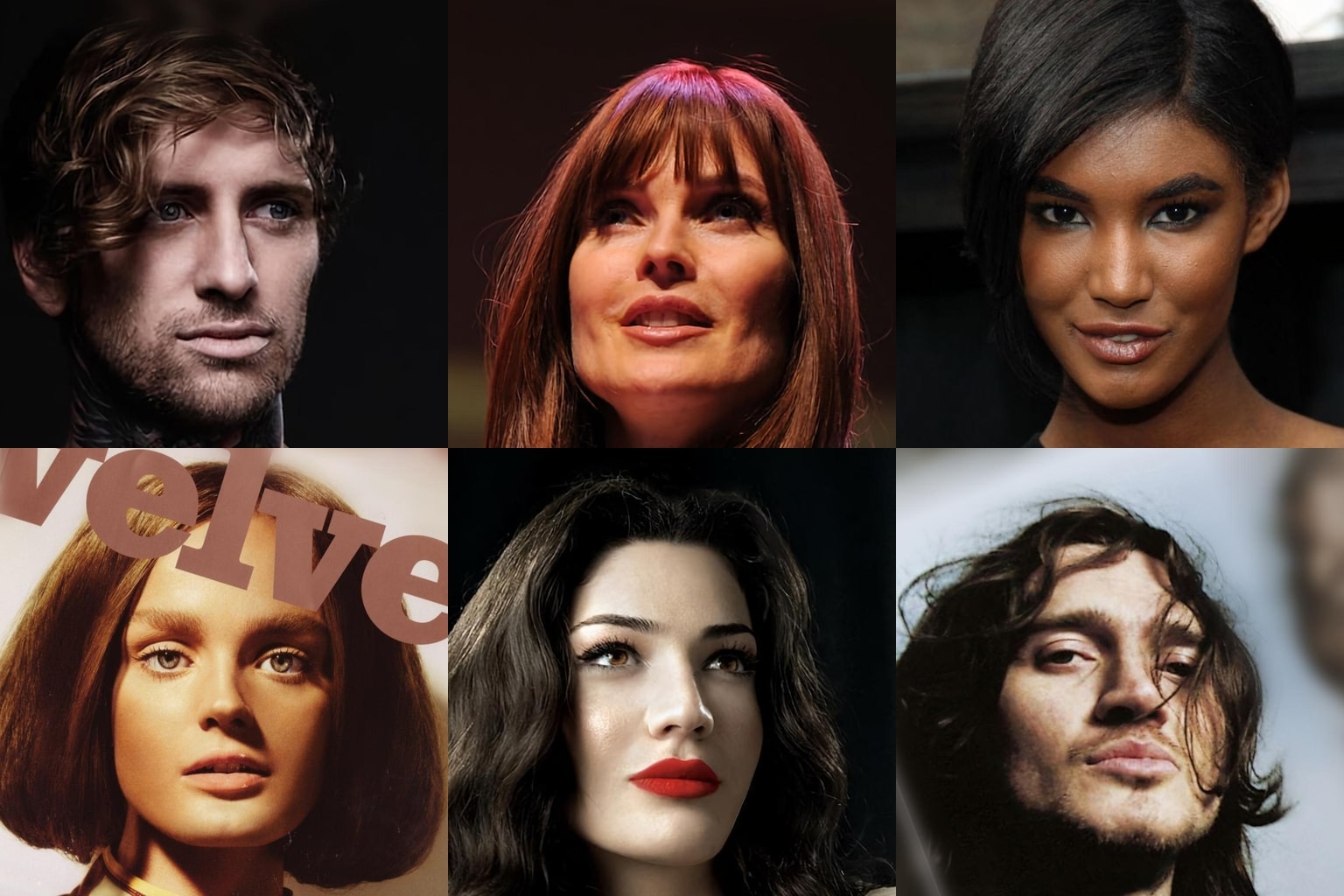}
                \caption{Nearest neighbors in the DINO space (FDD)}\label{fig:neighbors_00002_dino}
        \end{subfigure}
        \caption{Samples from our user study on feature space neighborhoods. For each reference image, we show its nearest neighbors in Inception and DINO feature spaces. Inception is biased towards objects (hat and microphone), while DINO can be perturbed by occluding objects (bottom). Therefore, Inception neighbors are not similar to the person, but simply wear similar hats.}
        \label{fig:survey-neighbors-00004}
\end{figure*}

\subsection{Investigating photorealism correlation}~\label{subsec:photorealism}
We expand with an analysis of the connection between FID/FDD relative to CelebA-HQ and the photorealism of image distributions. We conduct a second user study to obtain opinion scores on photorealism based on 10 images per category. The results of FID and FDD (Table~\ref{table:study-photorealism}) are closely related on the different sets. For FFHQ and truncated StyleGAN2~\cite{karras2020analyzing} images, the distances match well with opinion scores, however, they diverge for untruncated StyleGAN2 and PGGAN~\cite{karras2018progressive}, indicating that participant opinions are strongly affected by visual artifacts while the distance metrics focus more on content distributions. This is even more observable with PGGAN FDD; as PGGAN is trained on CelebA-HQ, its synthetic-image distribution matches better with it and leads to a low FDD despite lower visual quality. This further supports the claim that the specialized FDD focuses on high-level abstract information.

\subsection{Deeper dive into feature space neighborhoods}
Finally, we narrow down to an image-level analysis of the feature spaces. We exploit local neighborhoods to analyze the feature-space landscapes. We select reference images and find their respective nearest neighbors in each of the two spaces. Sample results are shown in Fig.~\ref{fig:survey-neighbors-00004}. We conduct a third user study where participants are asked to select which feature space induces neighbor images that are more similar to the reference (Table~\ref{table:study-similarity}). The Inception space is by a large margin better for Stanford Cars images, as expected. For cats or random CelebA-HQ images, the Inception space is more in accordance with human perception. However, when asking which neighbor set contains \textit{people} who are more similar to the reference \textit{person}, the DINO space correlates closer to the user choices for images with accessories. We note, however, that these aggregated results hide part of the analysis. Depicted in Fig.~\ref{fig:survey-neighbors-00004}, we observe that, indeed, Inception is excessively biased towards focusing on objects rather than faces. But for DINO, the lack of such bias did not guarantee the desired face similarity results, as seen in the bottom row. The specialization of DINO features on faces makes them untrained for other objects, which risk becoming similar to an adversarial attack.

\begin{table}[t]
    \centering
    \caption{User study rating the photorealism of images from various sources (1-5 score), and the corresponding \textit{rescaled} Fr\'echet distances relative to CelebA-HQ. $^*$PGGAN is trained on CelebA-HQ, while the other models are trained on FFHQ.}
    \begin{NiceTabular}{l|c|c|c|c}
    \toprule
    Image source distribution    & $\mu$ & $\sigma$ & FID & FDD \\
    \midrule
    FFHQ dataset samples         & 4.12 & 1.10 & 0.99 & 1.02 \\
    StyleGAN2 (0.7 truncated)    & 4.03 & 1.13 & 1.20 & 1.23 \\
    StyleGAN2 (untruncated)      & 3.19 & 1.44 & 1.09 & 0.94 \\
    PGGAN$^*$ dataset samples        & 1.93 & 1.11 & 0.83 & 0.09 \\
    \bottomrule
    \end{NiceTabular}
    \label{table:study-photorealism}
\end{table}

\section{Conclusion and Key Take-aways}~\label{sec:takeaways}

\vspace{-0.35cm}

We analyze an open question on feature-space distance, particularly, the effects on Fr\'echet distance, and neighborhoods, of specializing the feature extractor to the facial domain. Our experiments and user studies support the following findings. 

\begin{table}[t]
    \centering
    \caption{User study selecting which of Inception or DINO nearest neighbors are most similar to the reference. Numbers reported are mean vote percentages. The first 4 rows are based on \textit{image} similarity, "P." refers to \textit{person} similarity.}
    \begin{NiceTabular}{c|l|c|c|c}
    \toprule
    & Subset & Inception & DINO & $\sigma$ \\
    \midrule
    \multirow{4}{*}{\rotatebox{90}{\centering Image Sim}}
    & CelebA-HQ (accessories)                      & 59 & 41 & 20 \\
    & CelebA-HQ (random)                           & 72 & 28 & 14 \\
    \cdashline{2-5}
    & AFHQv2-Cats~\cite{choi2020stargan}          & 69 & 31 & 29 \\
    & Stanford Cars~\cite{krause2013collecting}   & 92 & 8 & 4 \\
    \hdashline
    \rotatebox{90}{\centering{P.}} & CelebA-HQ (accessories) & 42 & 58 & 24 \\
    \bottomrule
    \end{NiceTabular}
    \vspace{-0.4cm}
    \label{table:study-similarity}
\end{table}

\textbf{(1) Specialists become better at abstraction.} Our experiments highlight that our specialized feature extractor can learn abstract concepts pertaining to faces. The generalist focuses more on fine-granularity features that can be exploited across tasks, making it more sensitive to spatially localized loss of information and less sensitive to global changes like an upside-down face, as shown in Fig.~\ref{fig:celebahq_fid_trained}. 

\textbf{(2) Feature distance does not equate to photorealism.} Fr\'echet distance measures statistics over dataset image features. This is affected both by photorealism and degradations, but also by the general content distribution across the dataset. When computing the Fr\'echet distance \textit{relative} to a base dataset, it is important to use a high-quality one and to ensure that the base dataset contains a fair representation of desirable content. We emphasize that the vanilla distance is a holistic \textit{image}-based distance rather than a \textit{face} or identity distance. 

\textbf{(3) Noticing can be easier than not noticing.} While we can train specialists for features relevant to a specialized domain, this does not guarantee their ability to dismiss all irrelevant information. Facing novel content in their input can act as adversarial attacks perturbing the specialized network (Fig.~\ref{fig:survey-neighbors-00004}).

\textbf{(4) The risk of smaller specialized datasets.} Modern networks are large and this can lead to rich representations emerging even in randomly initialized ones. As the lottery ticket hypothesis~\cite{frankle2019lottery} hints, multiple paths lead to the final representation, enough for coincidental features to appear. Training improves this representation making it more practical. Particularly, training over a massive dataset such as ImageNet constrains the behavior of the feature extractor across its many paths. This advantage can be lost when training a large specialist network on smaller domain-specific datasets, possibly leading to the weakness described in (3).

Our findings fill a gap in the literature, highlighting the trade-offs between general and specialized feature extractors. One avenue for future research is hybrid training, preserving well constrained extractors with low-granularity features and robustness, while learning abstract specialized features. 

\newpage
\bibliographystyle{IEEEtran}
\bibliography{refs}

\end{document}


\title{- Supplementary Material - \\Exploring the Specialization of Fr\'echet Distance on Facial Images}

\author{Doruk Cetin$^{*\ddagger}$ \qquad Benedikt Schesch$^{*\dagger}$ \qquad Petar Stamenkovic$^{\dagger}$ \\ Niko Benjamin Huber$^{\ddagger}$ \qquad Fabio Z\"und$^{\dagger}$ \qquad Majed El Helou$^{\dagger}$}

\maketitle
\begin{abstract}
We present further details and illustrations that are omitted from the main manuscript for conciseness. We also provide an additional experiment that supports our claims regarding randomly initialized feature extractor networks. 
\end{abstract}

\noindent\let\thefootnote\relax\footnotetext{$^*$ Equal contribution, second author was an intern at MTC.}

\section{Datasets and image perturbations} \label{sec:datasets}

\begin{figure}[t]
        \centering
        \begin{subfigure}{.29\linewidth}
                \includegraphics[width=\linewidth]{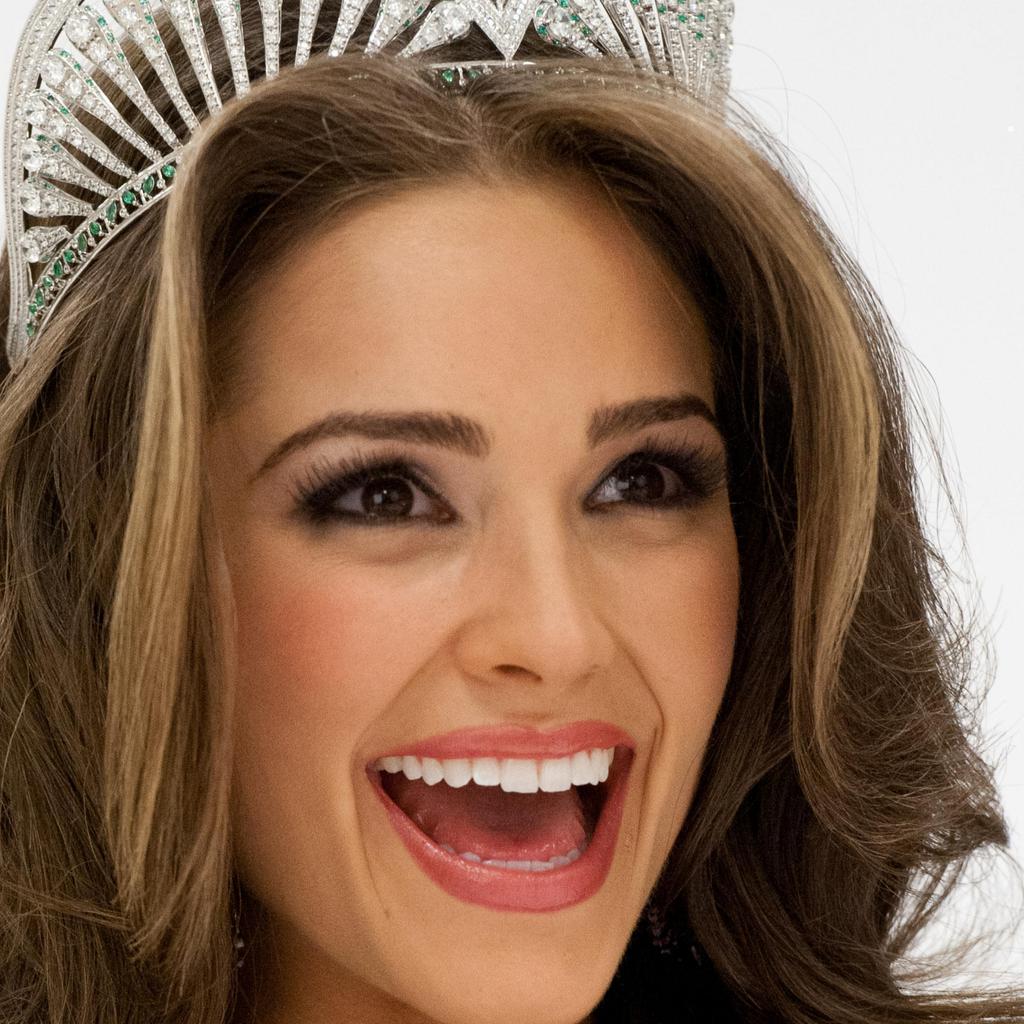}
                \caption{CelebAHQ}\label{fig:celebahq}
        \end{subfigure}
        \begin{subfigure}{.29\linewidth}
                \includegraphics[width=\linewidth]{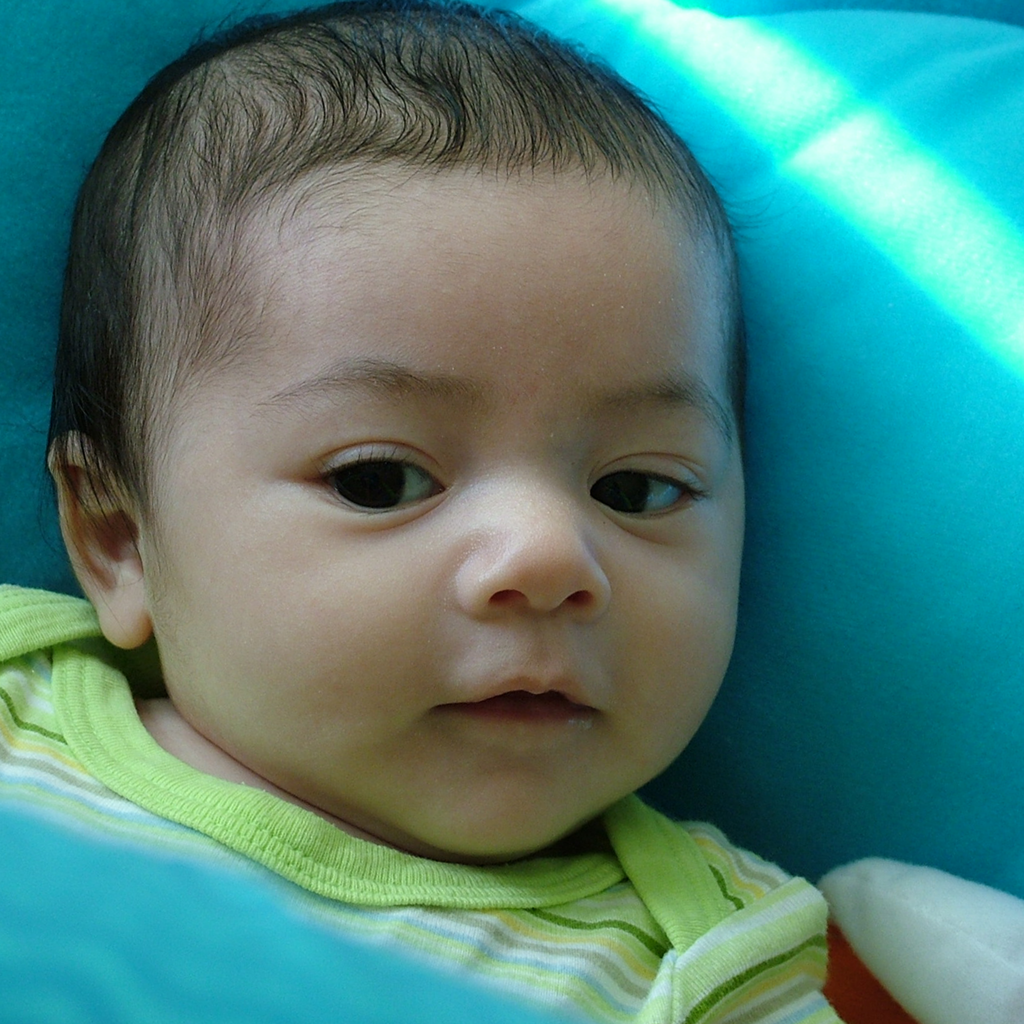}
                \caption{FFHQ}\label{fig:ffhq}
        \end{subfigure}
        \begin{subfigure}{.29\linewidth}
                \includegraphics[width=\linewidth]{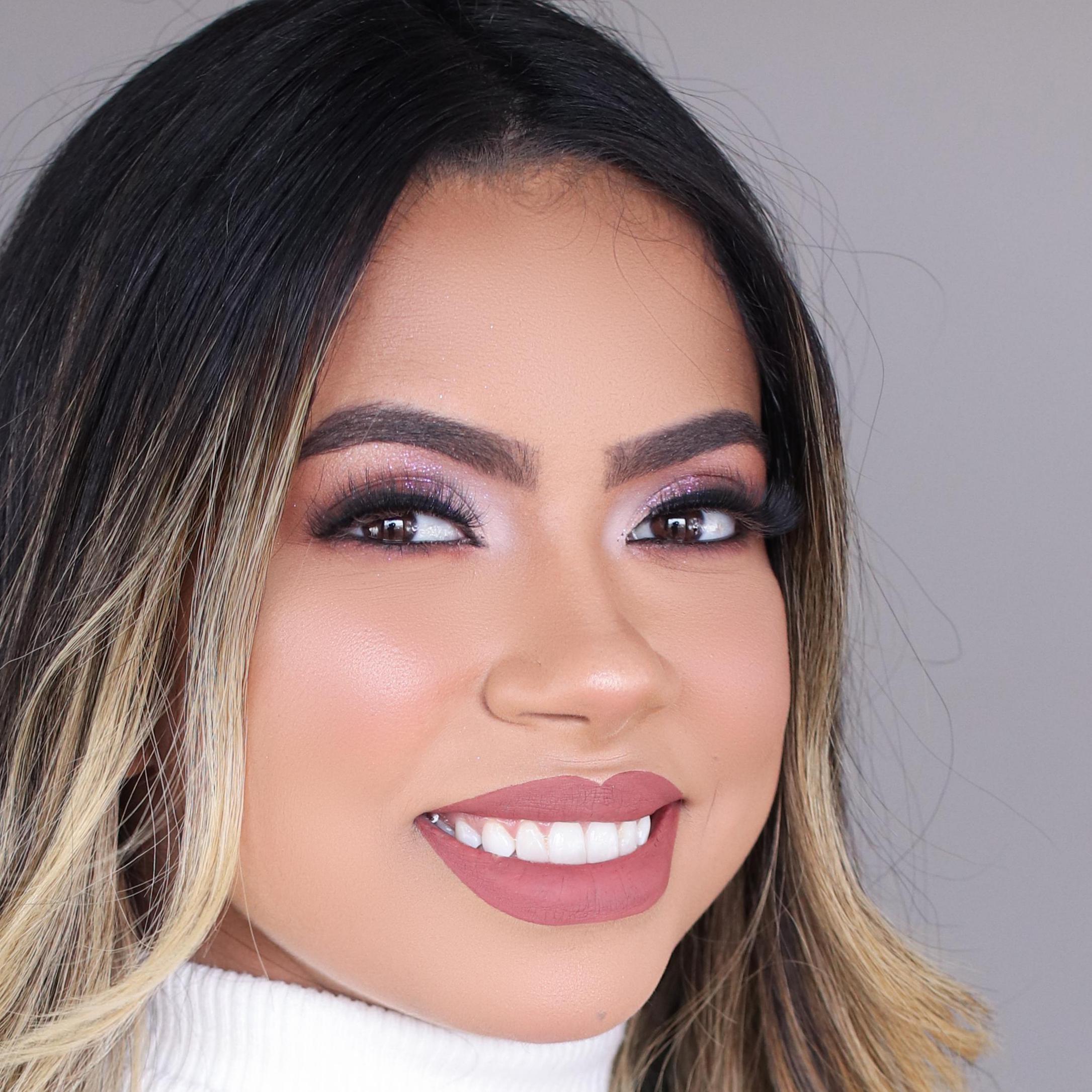}
                \caption{Faces}\label{fig:faces}
        \end{subfigure}
        \begin{subfigure}{.29\linewidth}
                \includegraphics[width=\linewidth]{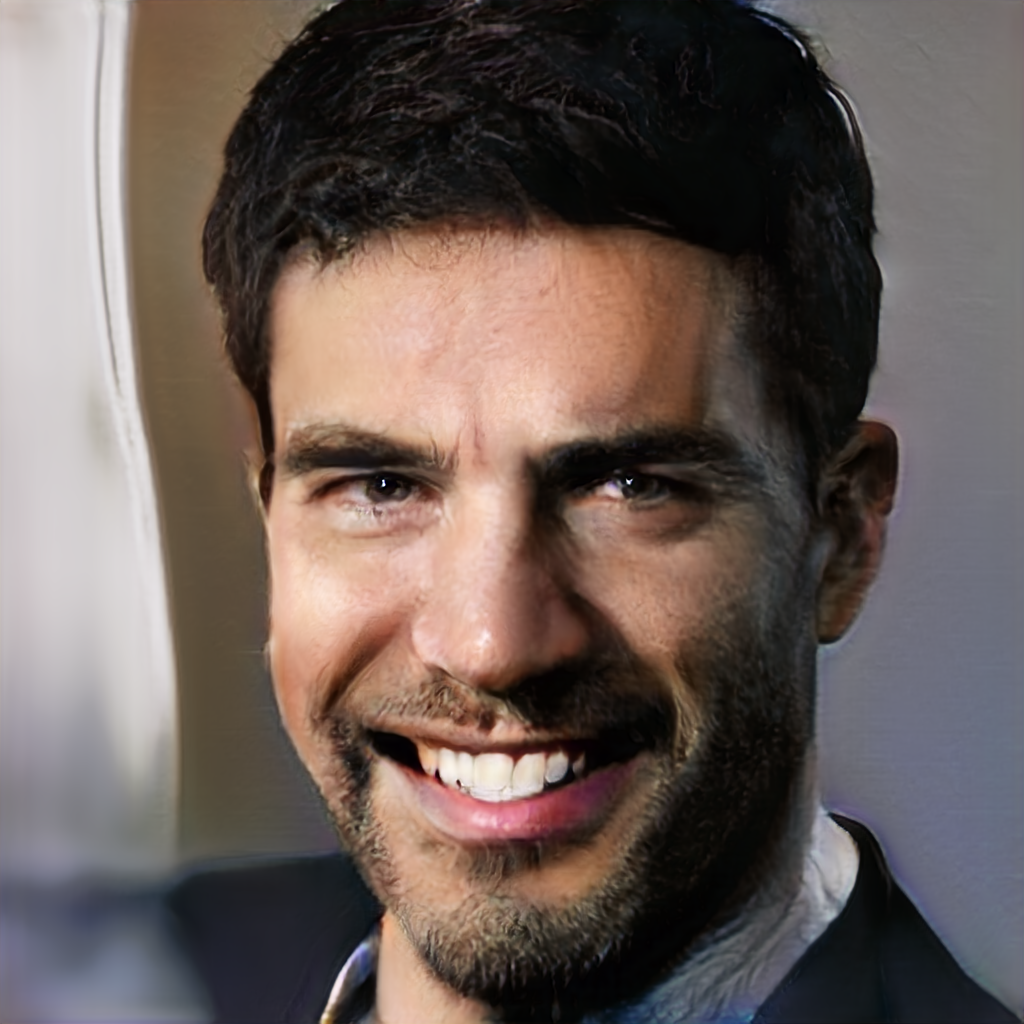}
                \caption{PGGAN}\label{fig:pggan}
        \end{subfigure}
        \begin{subfigure}{.29\linewidth}
                \includegraphics[width=\linewidth]{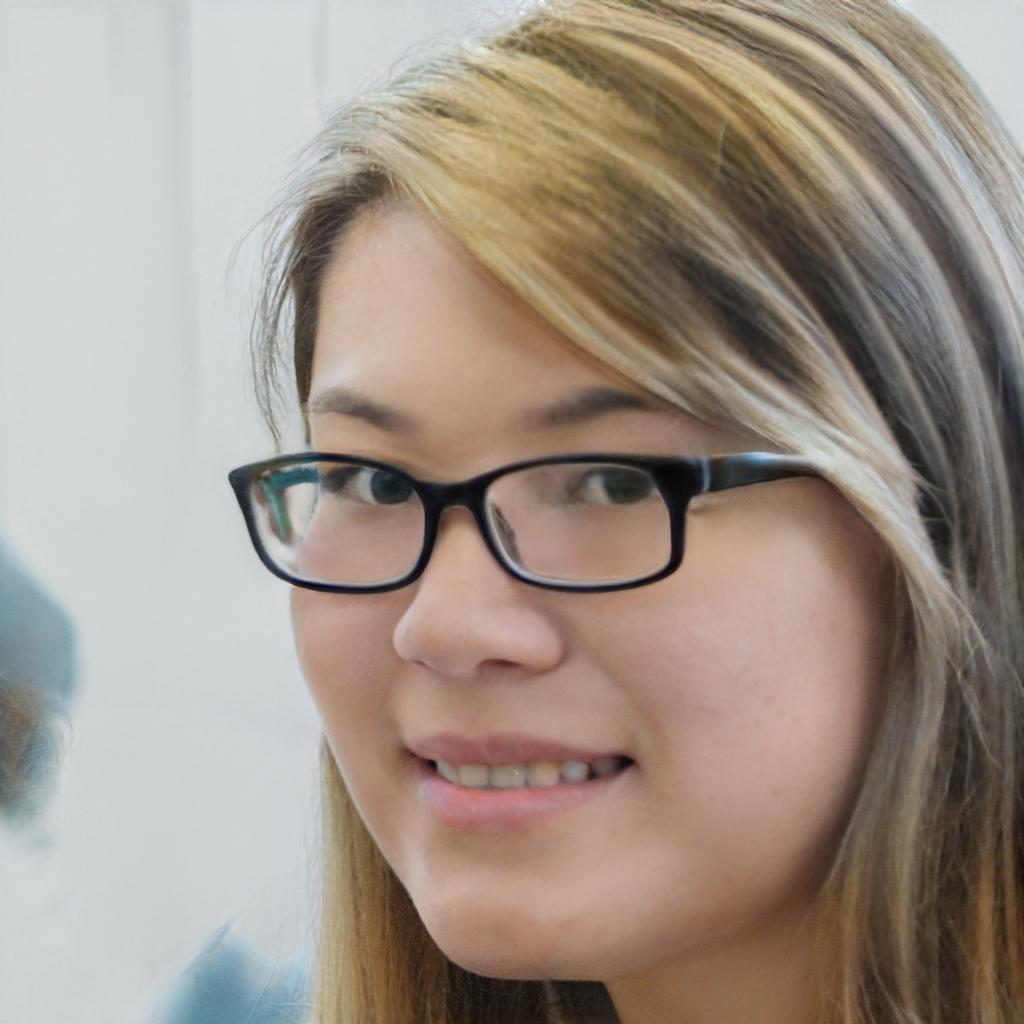}
                \caption{StyleGAN2}\label{fig:stylegan2_ffhq_1_0}
        \end{subfigure}
        \begin{subfigure}{.29\linewidth}
                \includegraphics[width=\linewidth]{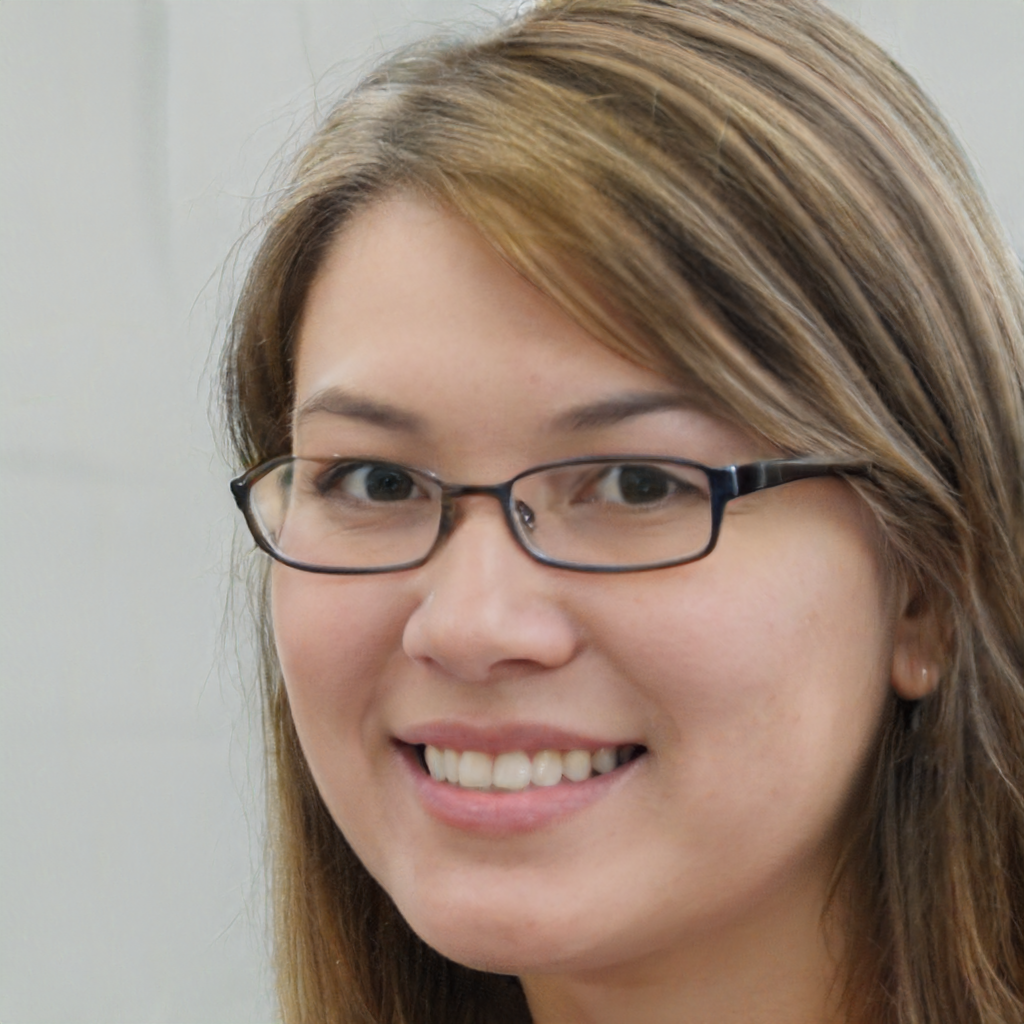}
                \caption{StyleGAN2$^*$}\label{fig:stylegan2_ffhq_07_0}
        \end{subfigure}
        \begin{subfigure}{.29\linewidth}
                \includegraphics[width=\linewidth]{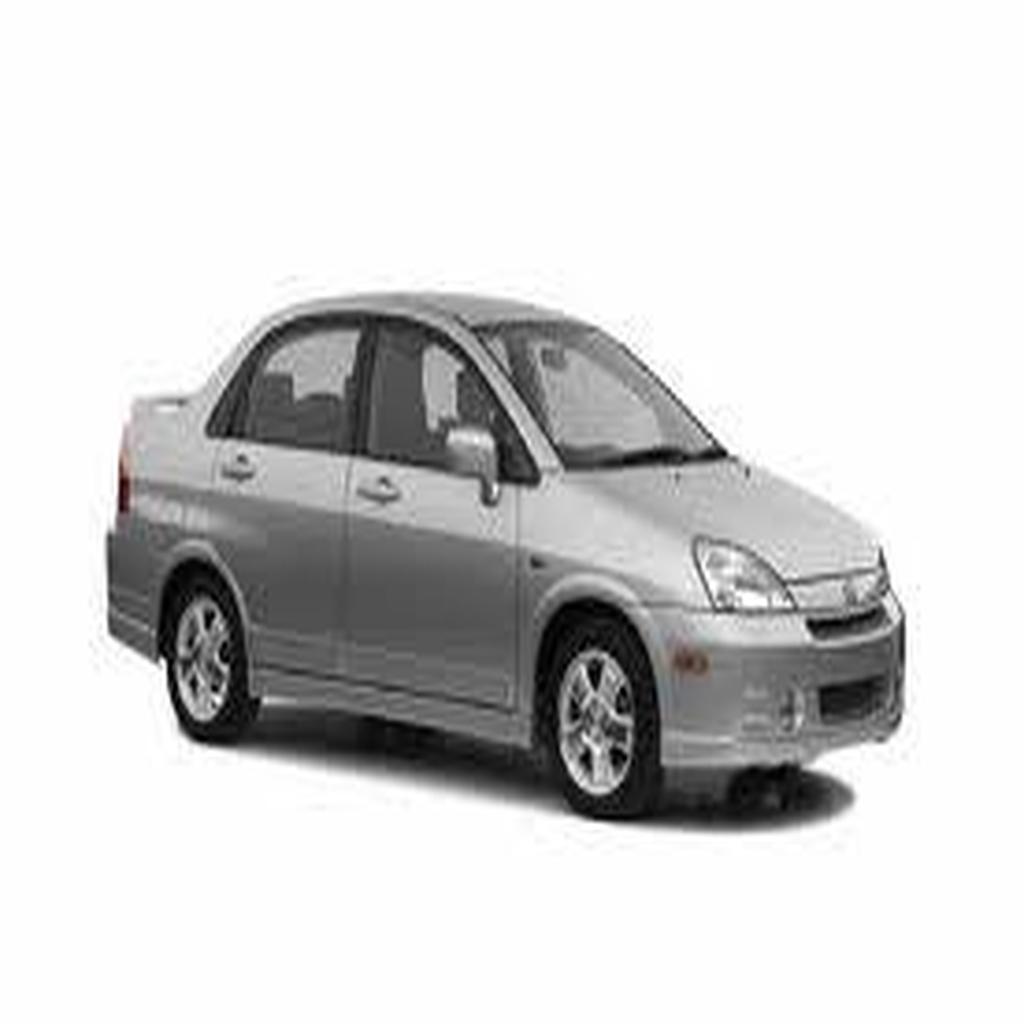}
                \caption{StanfordCars}\label{fig:stanfordcars_test_1k}
        \end{subfigure}
        \begin{subfigure}{.29\linewidth}
                \includegraphics[width=\linewidth]{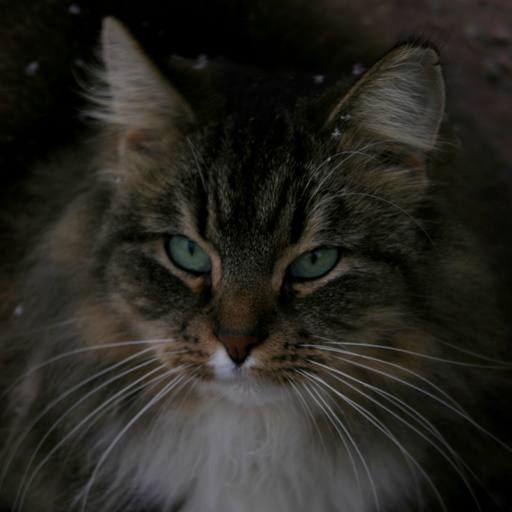}
                \caption{AFHQv2-Cats}\label{fig:afhq_v2_cat}
        \end{subfigure}
        \caption{\textbf{Dataset samples.} The first row illustrates the human faces datasets we use, whereas the second and third rows respectively contain samples from the datasets of synthetic face images and real non-human-face images. StyleGAN2$^*$ denotes the sample obtained with 0.7 truncation.}
        \label{fig:dataset-samples}
\end{figure}

\begin{figure}[t]
        \centering
        \begin{subfigure}{.29\linewidth}
                \includegraphics[width=\linewidth]{IMAGES_SUPP/samples/celebahq.png}
                \caption{Original}\label{fig:celebahq_original}
        \end{subfigure}
        \begin{subfigure}{.29\linewidth}
                \includegraphics[width=\linewidth]{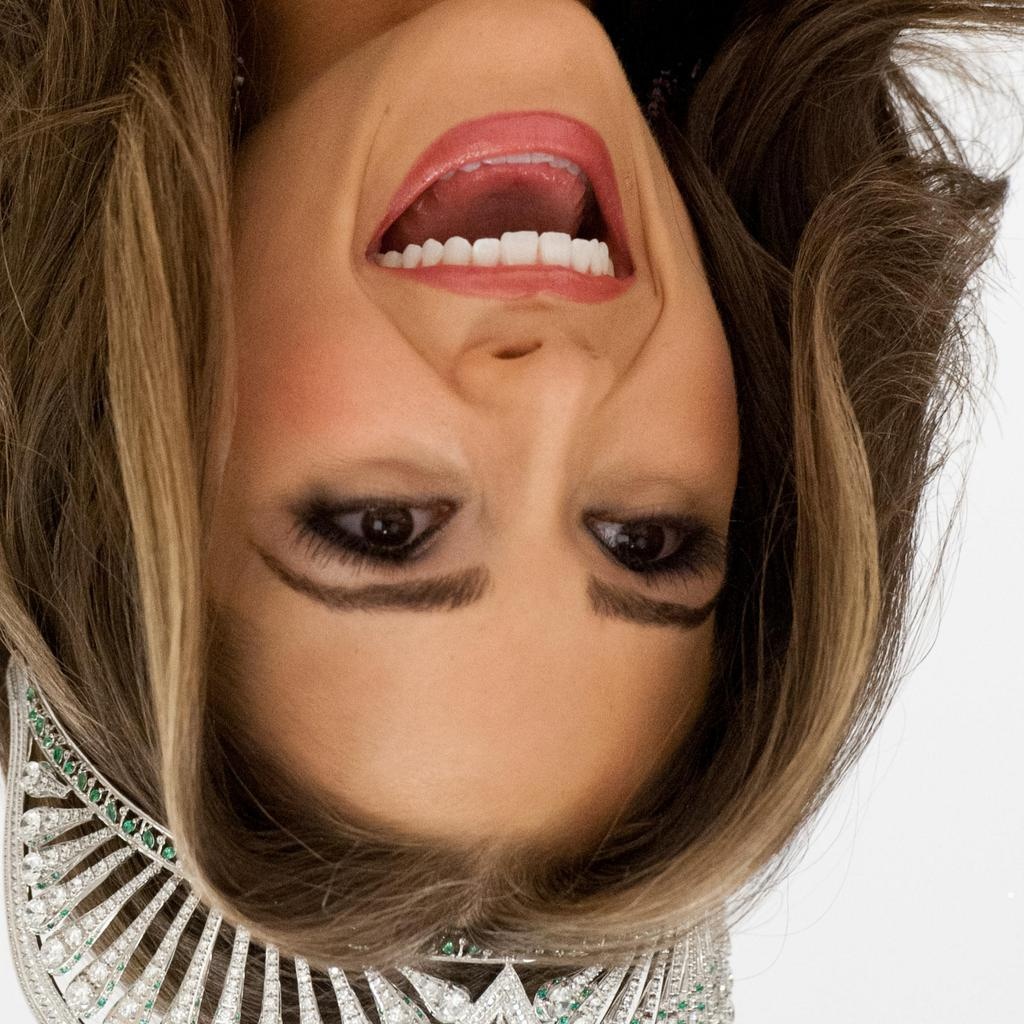}
                \caption{VerticalFlip}\label{fig:celebahq_vflip}
        \end{subfigure}
        \begin{subfigure}{.29\linewidth}
                \includegraphics[width=\linewidth]{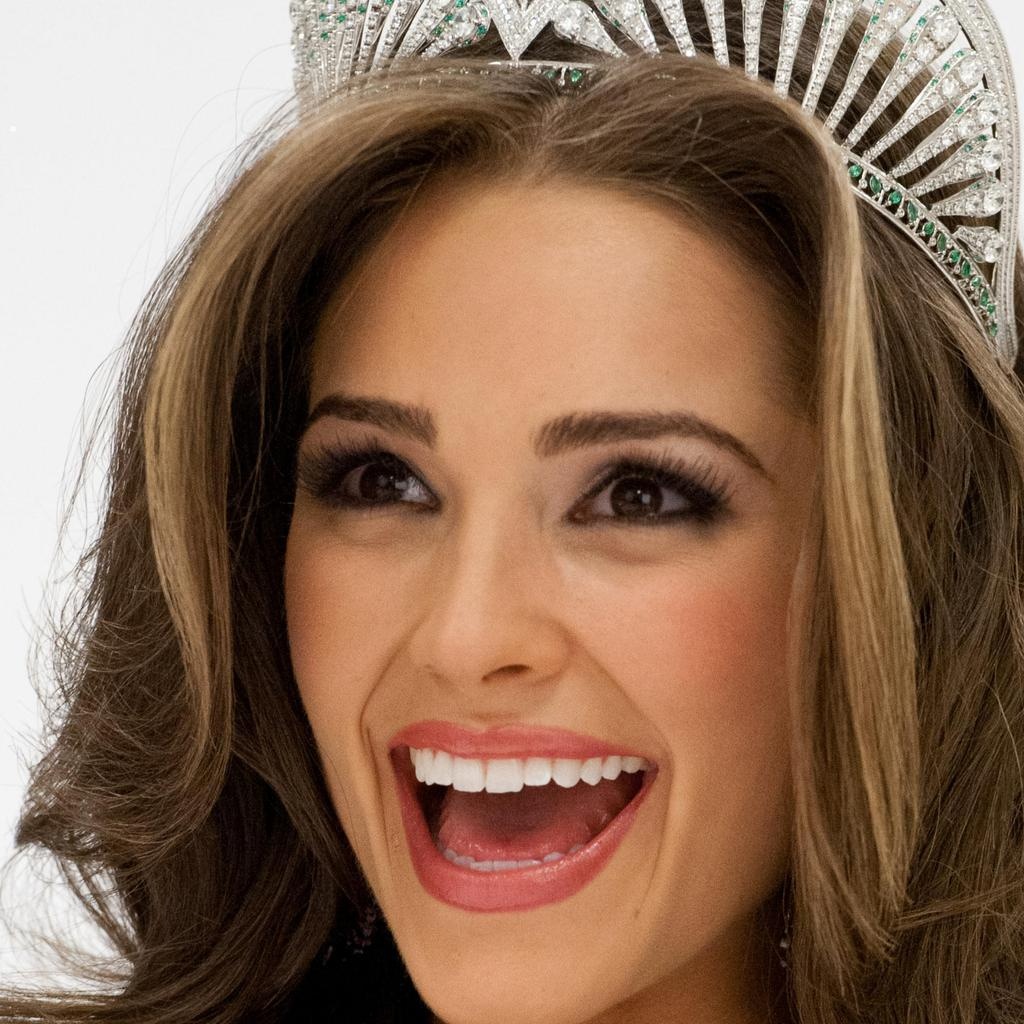}
                \caption{HorizontalFlip}\label{fig:celebahq_hflip}
        \end{subfigure}
        \begin{subfigure}{.29\linewidth}
                \includegraphics[width=\linewidth]{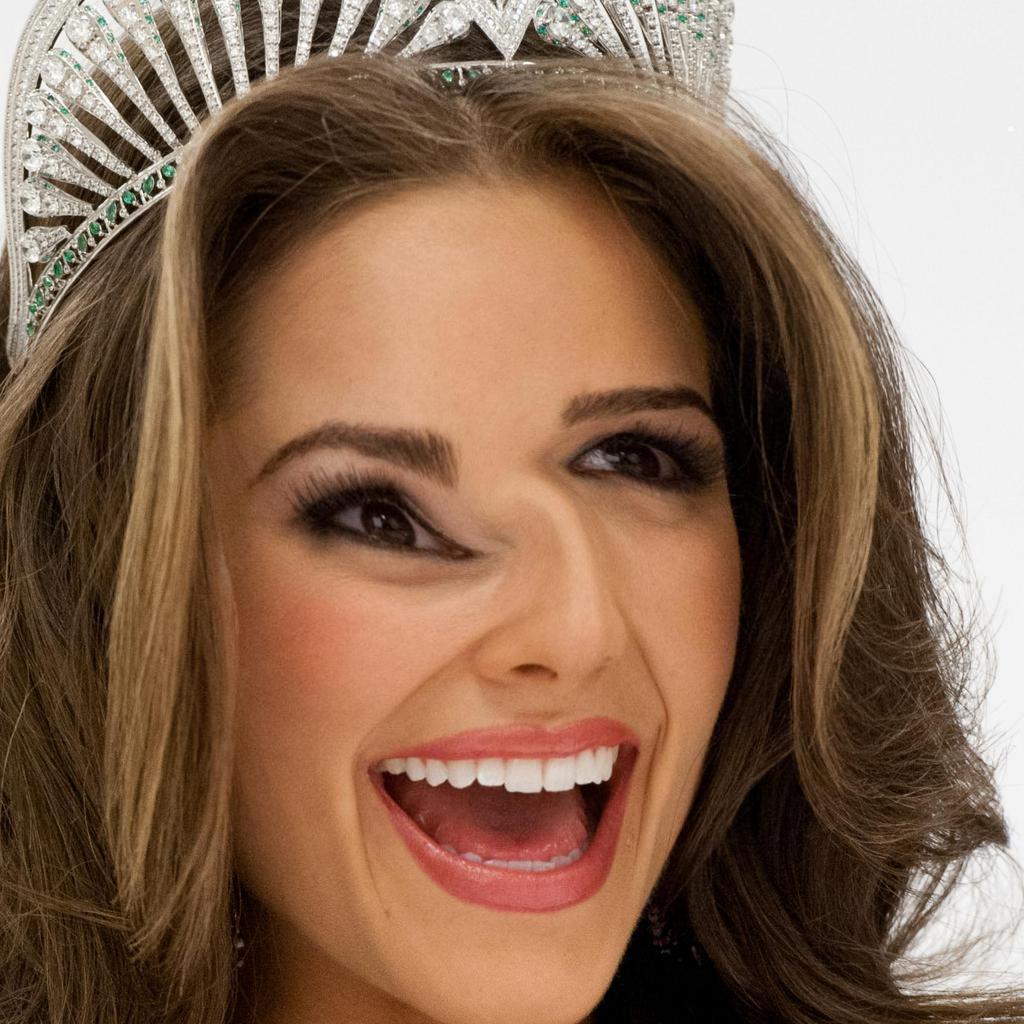}
                \caption{Swirl}\label{fig:celebahq_swirl}
        \end{subfigure}
        \begin{subfigure}{.29\linewidth}
                \includegraphics[width=\linewidth]{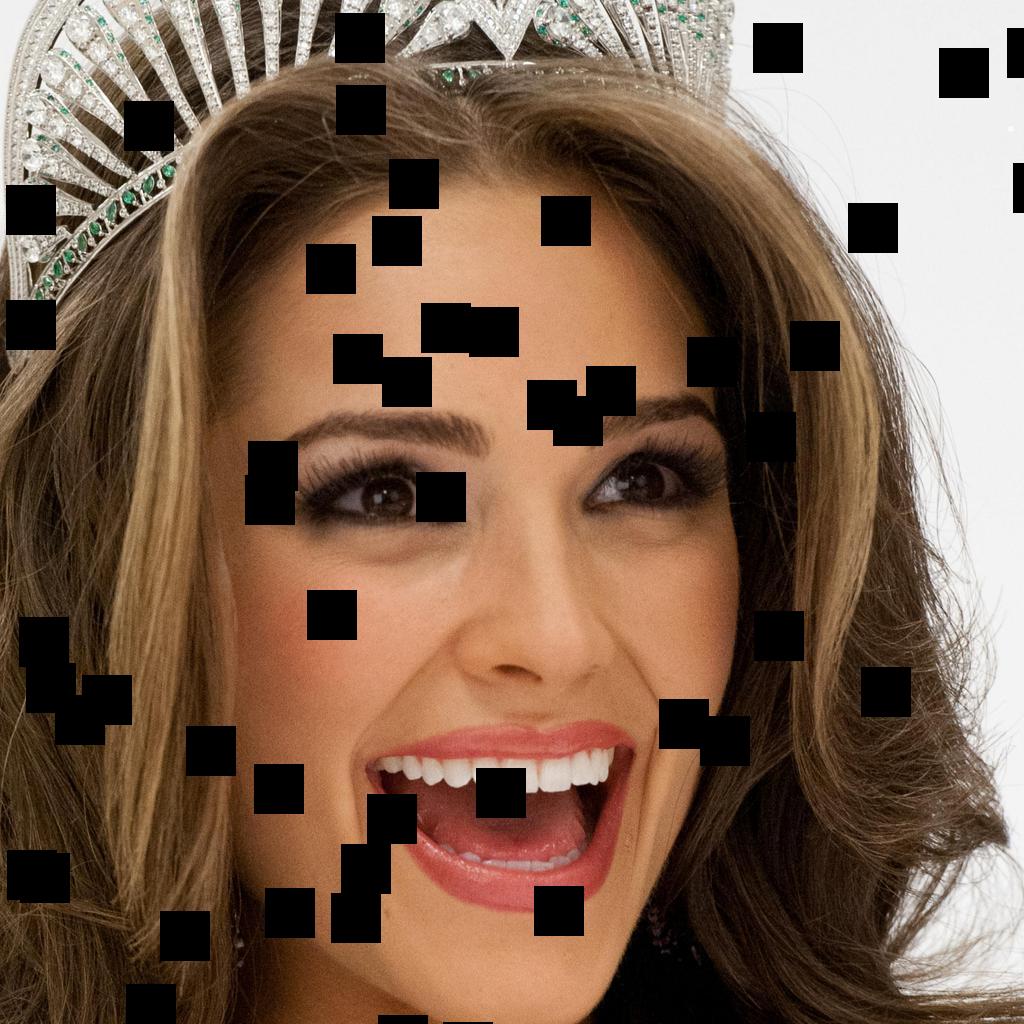}
                \caption{RandomErase}\label{fig:celebahq_randomerase}
        \end{subfigure}
        \begin{subfigure}{.29\linewidth}
                \includegraphics[width=\linewidth]{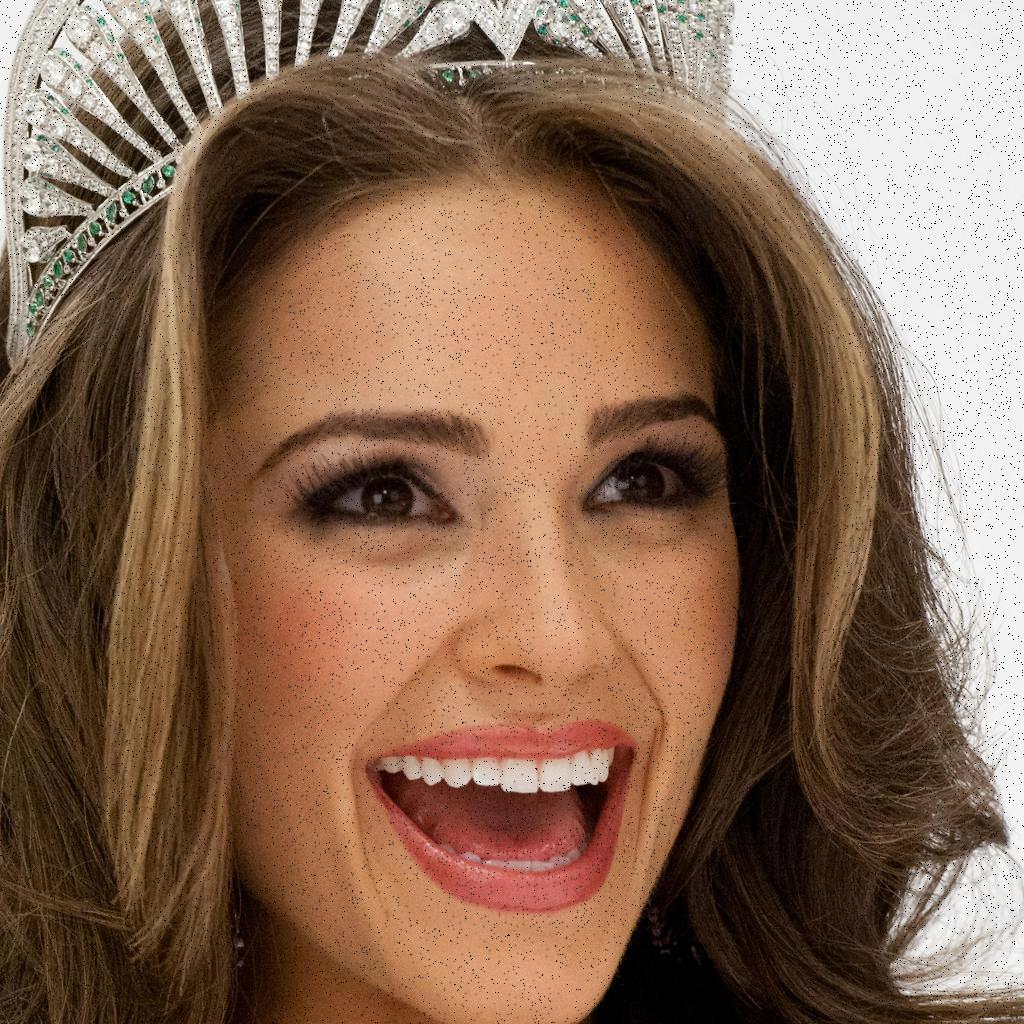}
                \caption{SaltPepperNoise}\label{fig:celebahq_saltpeppernoise}
        \end{subfigure}
        \begin{subfigure}{.29\linewidth}
                \includegraphics[width=\linewidth]{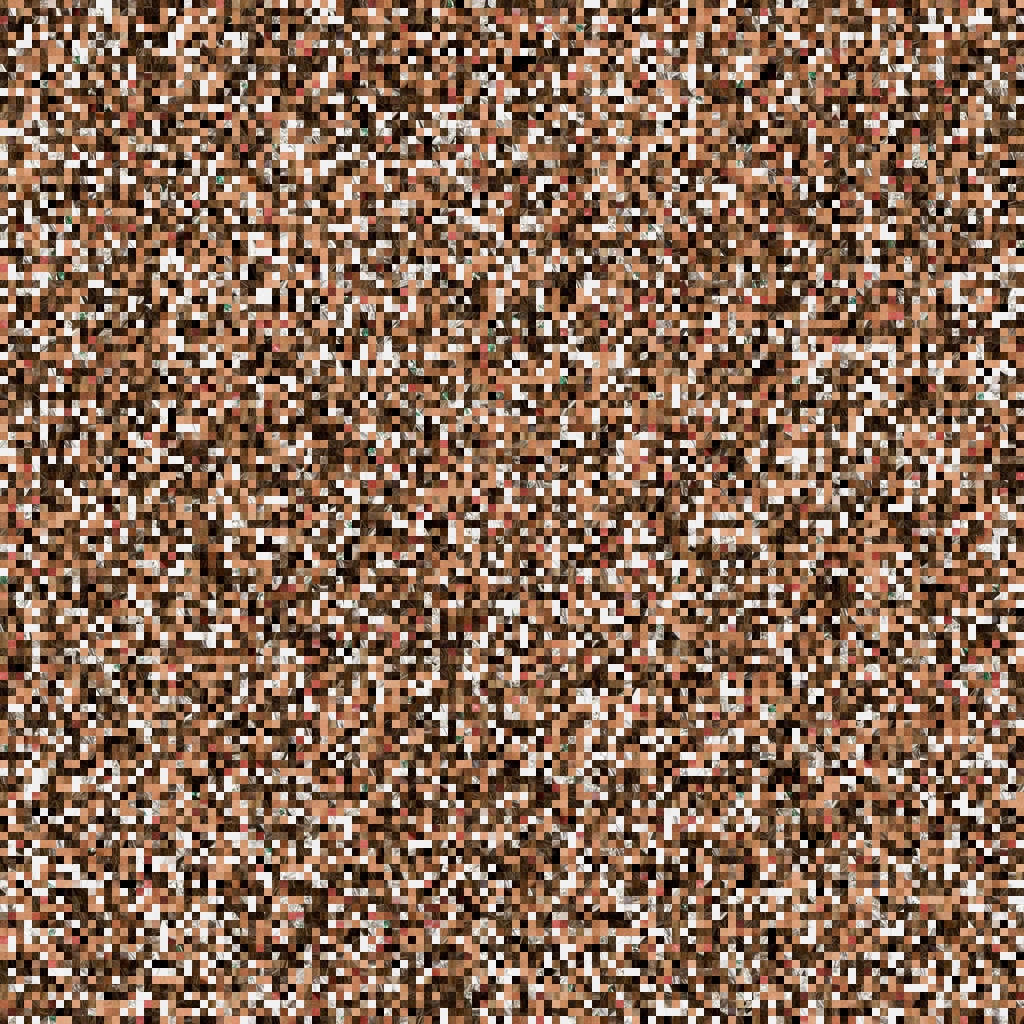}
                \caption{Puzzle8}\label{fig:celebahq_puzzle_8}
        \end{subfigure}
        \begin{subfigure}{.29\linewidth}
                \includegraphics[width=\linewidth]{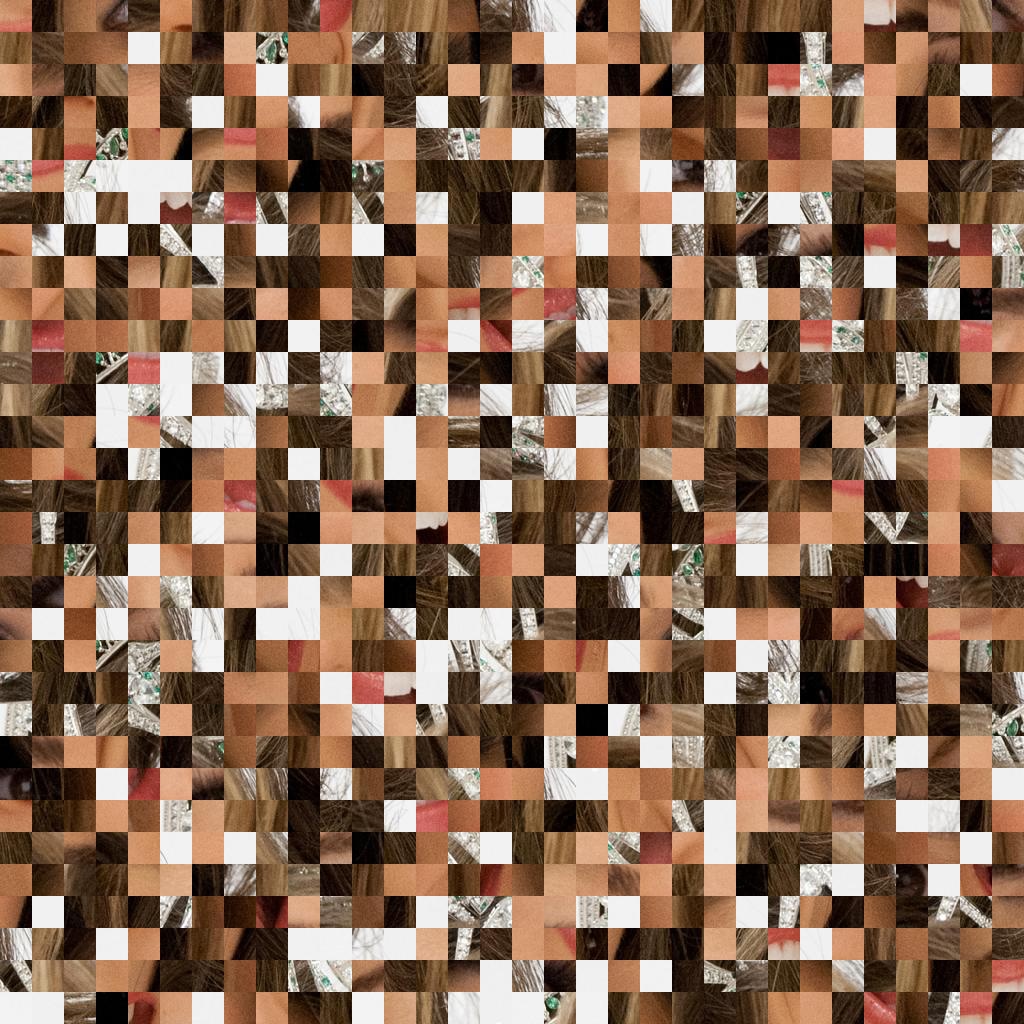}
                \caption{Puzzle32}\label{fig:celebahq_puzzle_32}
        \end{subfigure}
        \caption{\textbf{Image perturbations.} We illustrate the different perturbations that we apply to images in our experimental evaluation. The original input image is shown in (a) and the remaining images illustrate each of the different types of degradation models applied to the original image.}
        \label{fig:image-perturbations}
\end{figure}

\begin{figure*}[t]
    \centering
    \includegraphics[width=\textwidth]{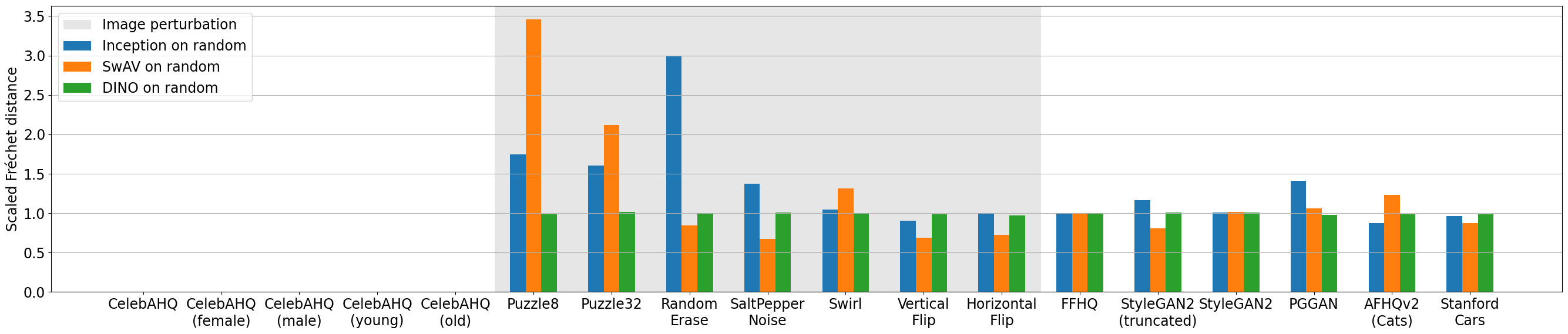}
    \caption{Rescaled Fr\'echet distance computed on randomly initialized Inception, SwAV and DINO features. We compute the Fr\'echet distances between the sets shown on the x-axis and CelebAHQ data (5k samples). For better readability, we rescale all values with a ratio that is \textit{fixed per method} and determined by an independent dataset. All values that are cut off for better readability are denoted next to their corresponding bars.}
    \label{fig:celebahq_fid_random}
\end{figure*}

We utilize several various datasets of real and synthetic images in our experiments. Image samples from the datasets are shown in Figure \ref{fig:dataset-samples} and the datasets are detailed below, grouped into three categories.

\textbf{Human face datasets:} CelebAHQ contains 30'000 aligned face images. FFHQ is an improvement over former in both quality and quantity, containing 70'000 images with less bias over attributes such as gender and age. Faces is a separate face dataset of 30'000 samples we collect, preprocessed in the same way as CelebAHQ and FFHQ. To promote better fairness, we also balance our dataset across six ethnicity groups, to alleviate concerns over ethnicity biases in such datasets as FFHQ.

\textbf{Synthetic datasets:} For PGGAN we construct a dataset over 10'000 samples generated by a network trained on CelebAHQ. For StyleGAN2, we similarly generate 10'000 samples but using a network trained on FFHQ instead. For the truncated version of the StyleGAN2, use a truncation value of $0.7$ and we keep the random seed fixed, thus samples in truncated and untruncated datasets match each other in terms of attributes of synthetic people they contain.

\textbf{Non-human-face datasets:} We also utilize two datasets that do not contain images of human faces. First is AFHQv2-Cats where we utilize all 5'558 cat images from AFHQv2 dataset, which contains face images of animals obtained through a similar preprocessing as in aforementioned face datasets. The second is Stanford Cars dataset, where we utilize the test set with 8'042 samples. It is the only dataset we use where the images do not have equal width and height of 1024 pixels and we resize all images to squares as preprocessing for the networks we use in our experiments.

\textbf{CelebAHQ attribute distribution:}
CelebAHQ provides annotations for several attributes for each image. We use two of those labels, namely "Young" and "Male", to sample class-specific data for our experiments and our survey. Table \ref{table:celebahq-stats} shows the distribution of the dataset with respect to these two labels, highlighting the biases of CelebAHQ. In our survey section on image correspondences we sample the images from CelebAHQ to construct image grids uniformly, such that the class distribution of all the images used in image grids match the percentages reported in Table \ref{table:celebahq-stats}.

\begin{table}[t]
    \centering
    \begin{tabular}{l|c|c|c|c}
    \toprule
    Label       & Male   & Female & Old    & Young  \\ \midrule
    Percentage  & $37\%$ & $63\%$ & $22\%$ & $78\%$ \\ 
    \bottomrule
    \end{tabular}
    \vspace{0.1cm}
    \caption{\textbf{CelebAHQ class distributions}. We show for reference the percentages of the different CelebAHQ classes used in our classification experiments.}
    \label{table:celebahq-stats}
\end{table}

We also explore the effects of several image perturbations on Fr\'echet distances in our experiments.
For VerticalFlip and HorizontalFlip we mirror the image in vertical and horizontal directions, respectively.
For Swirl, we apply a swirl transformation with strength of 2 and a radius of 400 pixels\footnote{\url{https://scikit-image.org/docs}}.
For RandomErase we black out 50 randomly selected patches of size $50x50$.
For SaltPepperNoise we switch up to 10'000 pixels ($0.95\%$ of a 1024x1024 image) each to black or white.
For Puzzle8 and Puzzle32 we respectively divide images into patches of size 8x8 and 32x32, then shuffle them randomly.
All image perturbations we use in our experiments are illustrated in Fig. \ref{fig:image-perturbations}.

\section{Scaling Fr\'echet distances} \label{sec:scaling}

\begin{table}[t]
    \centering
    \begin{tabular}{l|c|c}
    \toprule
    Model       & $\mu$ & $\sigma$ \\
    \midrule
    Inception (ImageNet) & 4.410         & 0.048 \\
    SwAV (ImageNet)      & 2.085         & 0.013 \\
    DINO (Faces)        & 1.368         & 0.059 \\
    \hdashline
    Inception (Random)   & 28533160.529  & 124058.421 \\
    SwAV (Random)        & 2737716.393   & 11397.390 \\
    DINO (Random)        & 15.803        & 0.005 \\
    \bottomrule
    \end{tabular}
    \caption{\textbf{Scaling factors per method.} We list for reference all the scaling factors used for improving the readability of our benchmarking results.}
    \label{table:scaling-factors}
\end{table}

There is no uniform scale over weights of different networks, trained or randomly initialized. Therefore, we normalize Fr\'echet distances for each model, diving it by a scaling factor calculated empirically, in order to be able to compare results from different networks more easily.

We randomly sample 5'000 samples each from CelebAHQ and FFHQ dataset and compare Fr\'echet distances over 10 different random seeds. Then, we use the average distance as the scaling factor for that network when we report distances elsewhere in the paper. Scaling factors used in the paper are listed in Table \ref{table:scaling-factors}, alongside the standard deviation of each run.

\section{Additional experimental results} \label{sec:results}
We show in Fig.~\ref{fig:celebahq_fid_random} similar benchmarking results as in our main manuscript, but with randomly initialized networks for each of the models. We observe that the results end up more or less uniform across any type of images that are outside of the CelebAHQ domain. We also note that the distances end up exploding in absolute values, and after rescaling we can see that the distances on the different CelebAHQ sets become so small that they are no longer readable. This can also be seen in Table~\ref{table:scaling-factors}. Therefore, while the randomly initialized networks can still extract a certain degree of relevant features, they are not practical to use, and as we discuss in our main text the training signals are indeed useful. 

\noindent \textbf{Acknowledgement:} Ringier, TX Group, NZZ, SRG, VSM, Viscom, and the ETH Zurich Foundation.